\documentclass[11pt]{article}

\usepackage[preprint]{acl}

\usepackage{times}
\usepackage{latexsym}
\usepackage[dvipsnames]{xcolor}
\usepackage[T1]{fontenc}

\usepackage[utf8]{inputenc}

\usepackage{microtype}

\usepackage{inconsolata}

\usepackage{graphicx}

\usepackage{multirow}
\usepackage{verbatim}
\usepackage{array}
\usepackage{algorithm}
\usepackage{algorithmic}
\usepackage{listings}
\usepackage{booktabs}
\usepackage{subcaption}
\usepackage{amsmath}
\usepackage{amssymb}
\usepackage[dvipsnames]{xcolor}

\newcolumntype{?}{!{\vrule width 2pt}}
\makeatletter
\newcommand{\thickhline}{%
    \noalign {\ifnum 0=`}\fi \hrule height 2pt
    \futurelet \reserved@a \@xhline
}
\newcolumntype{"}{@{\hskip\tabcolsep\vrule width 1pt\hskip\tabcolsep}}

\newcolumntype{?}{!{\vrule width 2pt}}
\makeatletter

\newcolumntype{"}{@{\hskip\tabcolsep\vrule width 1pt\hskip\tabcolsep}}

\title{TabEmb: Joint Semantic-Structure Embedding for Table Annotation \thanks{Published in the Proceedings of the 64th Annual Meeting of the Association for Computational Linguistics (ACL 2026).}}

\author{Ehsan Hoseinzade \\
  School of Computing Science \\
  Simon Fraser University \\
  Burnaby, Canada \\
  \texttt{ehoseinz@sfu.ca} \\\And
  Ke Wang \\
  School of Computing Science \\
  Simon Fraser University \\
  Burnaby, Canada \\
  \texttt{wangk@cs.sfu.ca} \\\And
  Anandharaju Durai Raju \\
  School of Computing Science \\
  Simon Fraser University \\
  Burnaby, Canada \\
  \texttt{aduraira@sfu.ca} \\}

\begin{document}
\maketitle
\begin{abstract}
Table annotation is crucial for making web and enterprise tables usable in downstream NLP applications. 
Unlike textual data where learning semantically rich token or sentence embeddings often suffice, tables are structured combinations of columns wherein useful representations must jointly capture column’s semantics \emph{and} the inter-column relationships. 
Existing models learn by linearizing the 2D table into a 1D token sequence and encoding it with pretrained language models (PLMs) such as BERT. However, this leads to limited semantic quality and weaker generalization to unseen or rare values compared to modern LLMs, and degraded structural modeling due to 2D-to-1D flattening and context-length constraints. We propose \textbf{TabEmb}, which directly targets these limitations by decoupling semantic encoding from structural modeling. An LLM first produces semantically rich embeddings for each column, and a graph-based module over columns then injects relationships into the embeddings, yielding joint semantic–structural representations for table annotation. Experiments show that TabEmb consistently outperforms strong baselines on different table annotation tasks. Source code and datasets are available at https://github.com/hoseinzadeehsan/TabEmb
\end{abstract}

\section{Introduction}

Tabular data is widespread and rapidly proliferating structured information on the web and in enterprise data lakes. Robust and scalable understanding of such data automatically is critical for downstream applications including data integration \cite{hai2023data}, data cleaning \cite{limaye2010annotating,kandel2011wrangler}, schema matching \cite{rahm2001survey}, data discovery \cite{fernandez2018seeping,fernandez2018aurum}, and  knowledge graph construction \cite{zhong2023comprehensive, steenwinckel2019csv2kg}.
A foundational step in this process is \emph{table annotation}, which assigns semantic labels to columns through Column Type Annotation (CTA) ~\cite{zhang2019sato,hul2019sherlock, suhara2022annotating,deng2020turl,hoseinzade2024graph,hoseinzadeefficient,hoseinzade2026ztab,zhang2024tablellama,korini2023column}, to relations between columns through Column Property Annotation (CPA) ~\cite{deng2020turl,korini2024column,suhara2022annotating,zhang2024tablellama}, and to entire tables through Table Type Annotation (TTA) ~\cite{kayali2024chorus,korini2023column,korini2024column}. 

While recent works explored LLM-based zero- and few-shot approaches for some of these tasks without task-specific training for minimal reliance on labeled data \cite{zhang2024tablellama, korini2023column, korini2024column, kayali2024chorus, zhang2024jellyfish}, they under-perform compared to supervised approaches on complex CTA/CPA/TTA datasets with large label spaces \cite{zhang2024tablellama, feuer2024archetype}. 
Within supervised table annotation, recent progress has been driven by improved \emph{column embedding} representations. Effective representations must satisfy two requirements: (i) \emph{high semantic fidelity}, capturing meaning of a column’s content, and (ii) \emph{explicit structural modeling}, capturing relationships among table's columns. From this perspective, supervised methods are broadly categorized into three regimes. 

\textit{Single-column approaches} \cite{chen2019colnet,hul2019sherlock, feuer2024archetype, sun2023reca} aim at learning semantics of \emph{individual} columns relying on their statistical features or via textual encoders, however, they ignore intra-table dependencies that are often essential for disambiguation.

\textit{Structure-augmented approaches} \cite{zhang2019sato, hoseinzade2024graph} address dependencies between columns, but typically as a post-hoc refinement over prediction outputs (logits) and serve only a "structure-blind prediction" rather than learning "structure-aware" embeddings. 

\begin{figure*}[t]
\centering

\begin{subfigure}{0.32\textwidth}
    \centering
    \includegraphics[width=\linewidth, height=4cm, keepaspectratio=false]{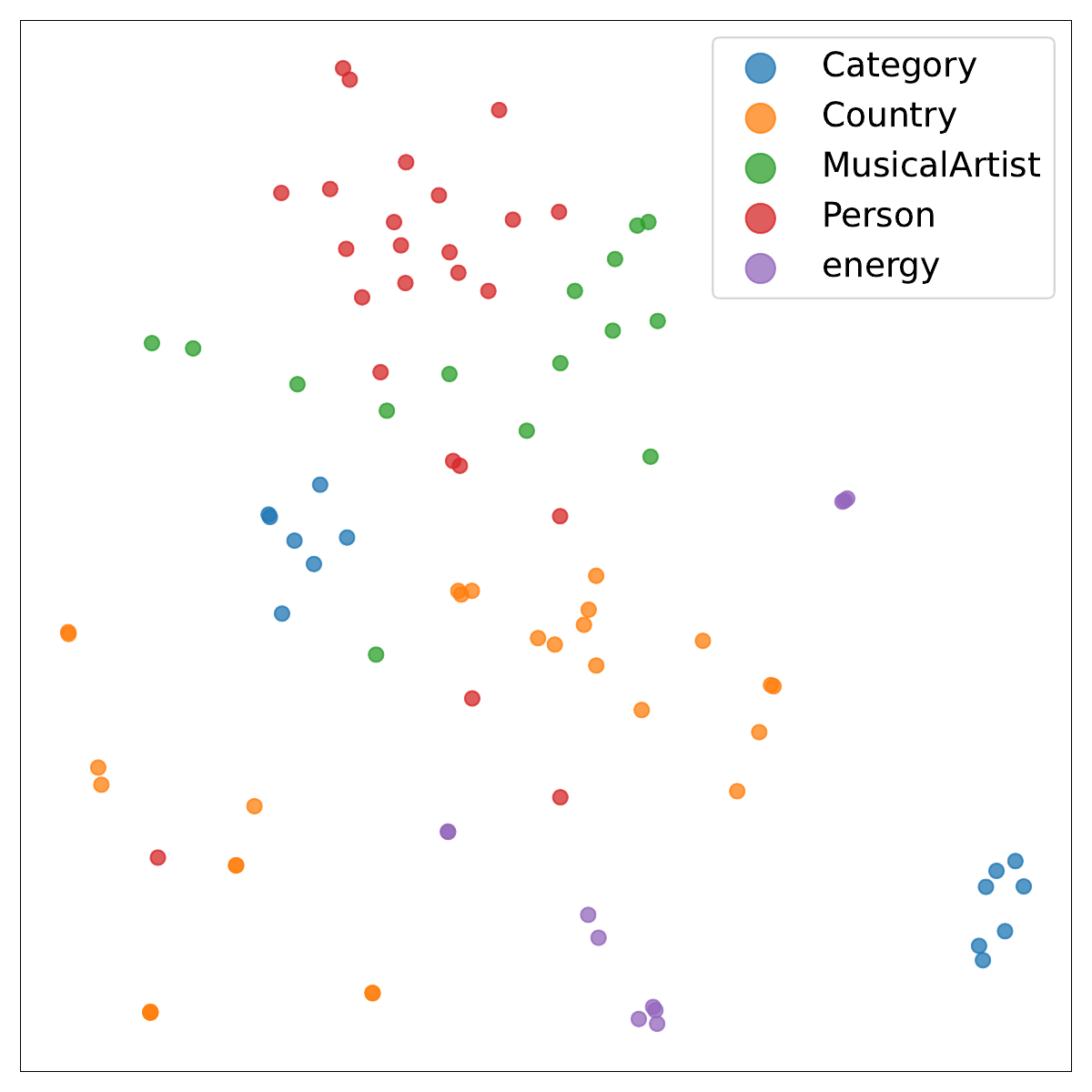}
    \caption{PLM (BERT)}
    \label{fig:embed-bert}
\end{subfigure}
\hfill
\begin{subfigure}{0.32\textwidth}
    \centering
    \includegraphics[width=\linewidth, height=4cm, keepaspectratio=false]{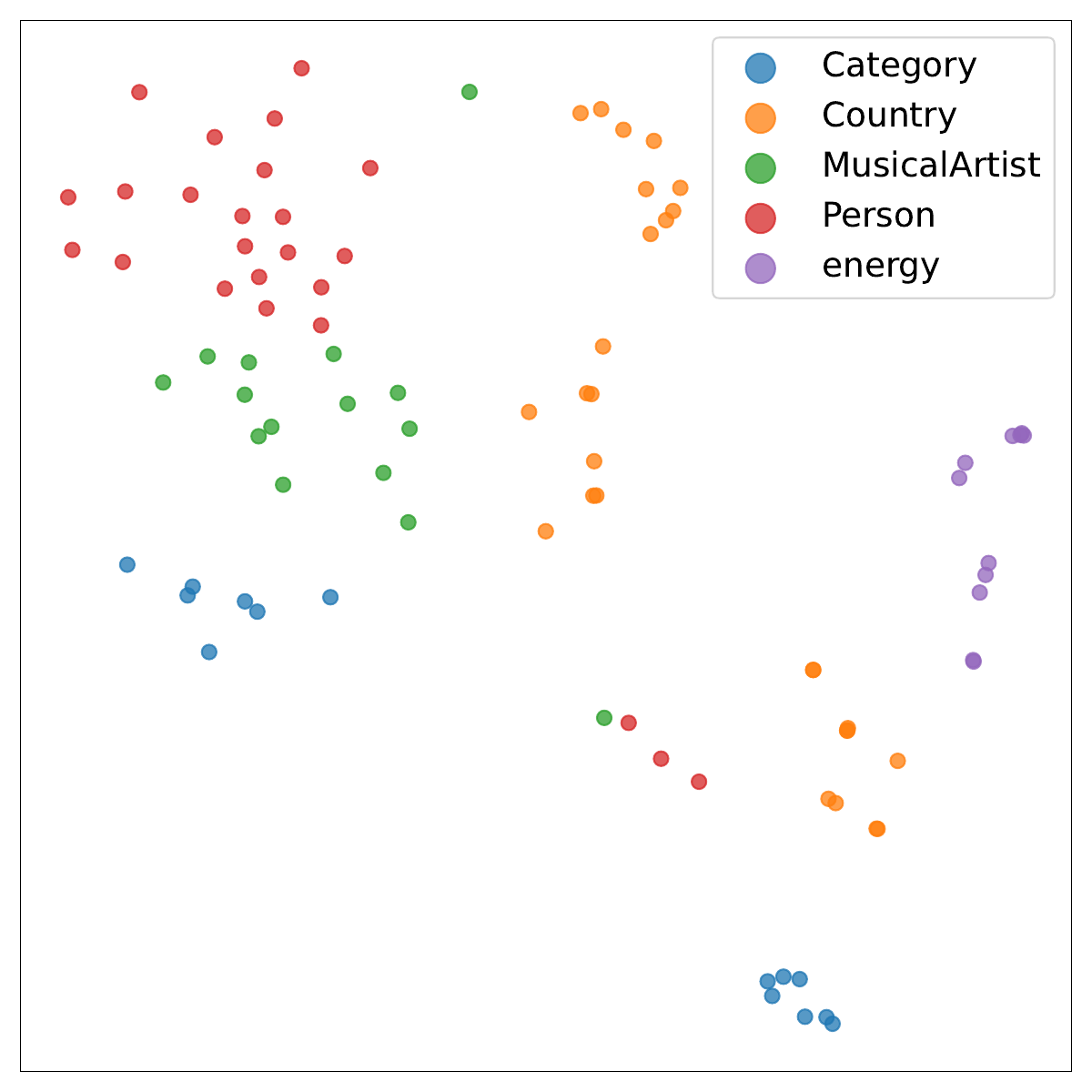}
    \caption{LLM (Mistral)}
    \label{fig:embed-llm}
\end{subfigure}
\hfill
\begin{subfigure}{0.32\textwidth}
    \centering
    \includegraphics[width=\linewidth, height=4cm, keepaspectratio=false]{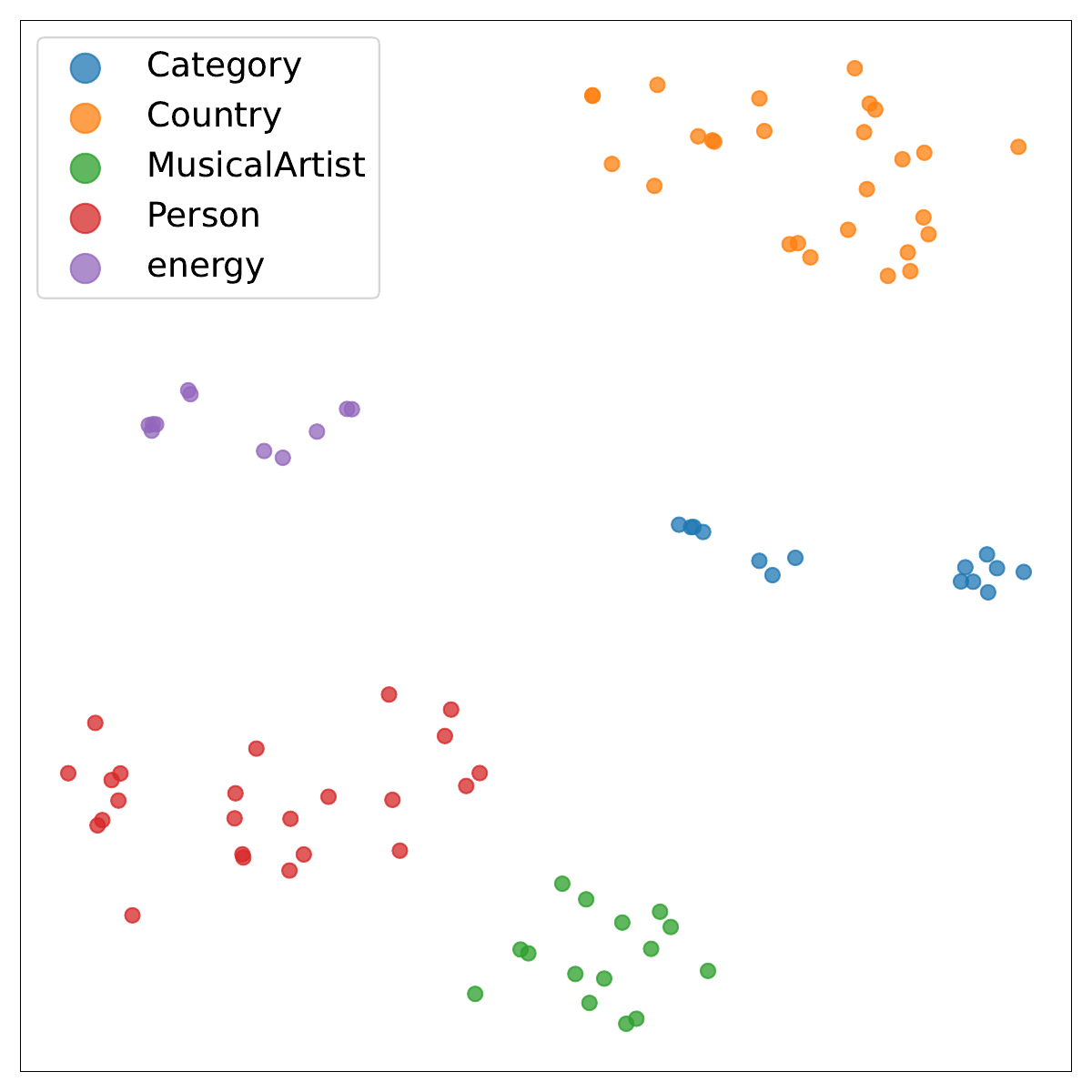}
    \caption{TabEmb (Mistral + Trainable GNN)}
    \label{fig:embed-gnn}
\end{subfigure}
\vspace{-2mm}
\caption{\textbf{The Impact of joint semantic-structure Modeling.} 
  t-SNE visualization of column embeddings from three paradigms: (Left) \textbf{PLM (BERT)}, (Middle) \textbf{LLM (Mistral)}, and (Right) \textbf{TabEmb (Mistral + Trainable GNN)} on the SOTAB\textsubscript{dbp} dataset. For readability, we visualize only a subset of classes.
  While BERT produces weakly separated clusters and the LLM improves semantic coherence, the LLM still exhibits significant overlap between ambiguous types (e.g., \textit{Person} vs. \textit{MusicalArtist}) due to its lack of structural context. 
  By explicitly modeling table structure via a GNN, TabEmb resolves this ambiguity, yielding the most discriminative embedding space.}

\label{fig:embedding-quality}
\end{figure*}

\textit{Joint modeling approaches} \cite{yin2020tabert,deng2020turl,iida2021tabbie,
wang2021tcn,suhara2022annotating,ding2025retrieve} attempt to jointly capture semantics and structure using pretrained language models (PLMs) like BERT by encoding entire tables. However, three limitations remain. First, they have \textbf{limited semantic coverage and world knowledge} compared to modern LLMs, leading to weaker generalization to unseen column values at test time and less well-separated clustering in the column embedding space
(Fig.~\ref{fig:embedding-quality}a). Second, linearizing tables from 2D to 1D sequences \textbf{destroys the native relational structure} and imposes strict input-length limits (e.g., 512 tokens for BERT), degrading performance on wide tables. Third, adapting PLMs to tabular structure typically requires \textbf{expensive PLM fine-tuning}.

\begin{figure*}[t]
\centering
\includegraphics[width=1\textwidth, height=4.0cm, keepaspectratio=false]{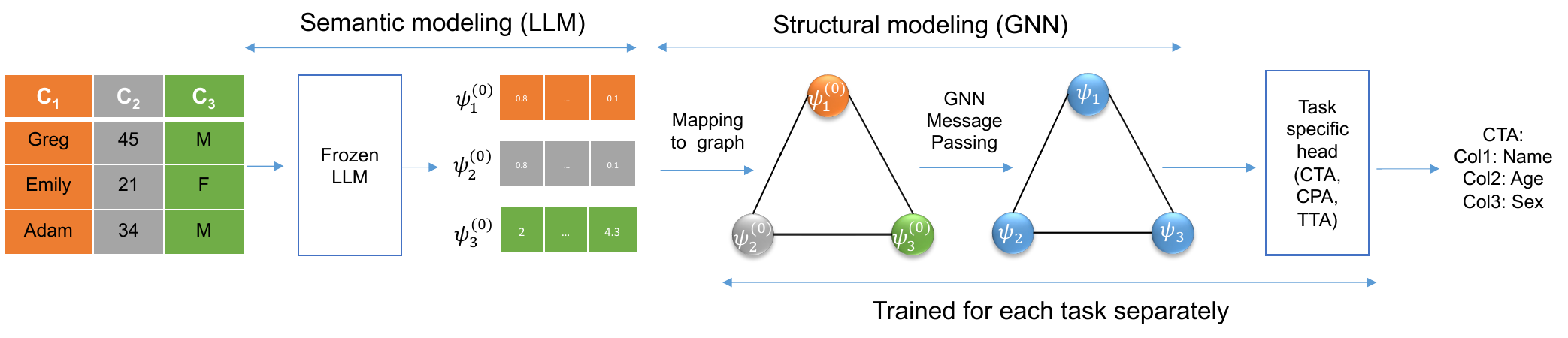} 
\vspace{-5mm}
\caption{Overview of TabEmb. A frozen LLM encodes each column into a semantic embedding, a GNN refines these via message passing over the column graph, and a lightweight task-specific head predicts labels, with the GNN and head trained separately for each task (CTA/CPA/TTA).}
\label{fig: overall TabEmb}
\end{figure*}

To address these challenges, we introduce \textbf{TabEmb} (Fig \ref{fig: overall TabEmb}), a modular framework for learning \emph{joint semantic-structural embeddings (structure-aware embeddings)} for supervised table annotation. TabEmb encodes each column independently using a frozen LLM, yielding semantically rich and well-separable embeddings (Fig.~\ref{fig:embedding-quality}b) that generalize better to unseen values and rare semantic types. It then uses a trainable graph neural network (GNN) to perform message passing over a fully-connected graph of columns, injecting intra-table structure directly at the \emph{embedding level}. Finally, the task-specific classification heads operate on these embeddings to support CTA/CPA/TTA. By sharing the frozen LLM backbone across tasks, and training the GNNs independently for the demands of CTA/CPA/TTA tasks, TabEmb combines LLM-level semantic quality with explicit structural modeling producing more discriminative embedding spaces (Fig.~\ref{fig:embedding-quality}c) while remaining computationally practical by avoiding expensive full-model fine-tuning. 

Our core contributions are as follows:
\begin{itemize}
    \item We introduce \textbf{TabEmb}, a modular framework that combines frozen LLM-based semantic encoding with trainable GNN-based structural modeling to learn high-quality semantic-structural column embeddings (Figure \ref{fig:embedding-quality}).
    \item We benchmark TabEmb across the CTA, CPA and TTA tasks, six datasets, and 8 baselines. TabEmb outperforms all prior methods in micro-F1 score, achieving average gains of 4.2 on CTA, 4.7 on CPA, and 5.6 on TTA over the strongest baseline for each task (Table \ref{tab:model_comparison}).
    \item TabEmb trains significantly faster than fine-tuned LLM baselines by keeping the LLM frozen and training only lightweight GNN and classification heads (Table~\ref{tab:time_summary}).
    
\end{itemize}

\vspace{5mm}

\section{Related Work}

Recent methods for table annotation can be grouped into four categories by \emph{what} they model:  


\textbf{Single-column models.}
These early approaches focus on modeling each column independently. 
ColNet \cite{chen2019colnet} uses DBpedia lookups to create training examples and learns a CNN over column values. Sherlock \cite{hul2019sherlock} combines hand-crafted statistical and textual features with a feed-forward network. RECA \cite{sun2023reca} leverages BERT with retrieved inter-table context to strengthen per-column representations. 
ArcheType \cite{feuer2024archetype} fine-tunes an open-source LLM on column-type labels. 
These methods emphasize semantic quality of columns' representation but ignore intra-table dependencies.

\textbf{Structure-augmented models.}
These approaches exploit relationships between columns given initial predictions. 
SATO \cite{zhang2019sato} extends Sherlock with a CRF over adjacent columns to enforce local dependencies. GAIT \cite{hoseinzade2024graph} applies a GNN over RECA’s column-level predictions to capture intra-table dependencies. 
In both cases, structure operates as a refinement layer \emph{on top of} prediction of another model, rather than being integrated into the embeddings.

\textbf{Joint semantic-structural models.} 
Joint models aim to encode both semantics of column content and table structure in a unified representation. HNN \cite{chen2019learning} models intra-column semantics, enhancing ColNet. With pretrained language models, several methods feed an entire table (or large portions of it) to a PLM like BERT-style encoders and fine-tune the model. TaBERT \cite{yin2020tabert} utilizes BERT as a base model to capture the table content features; TURL \cite{deng2020turl} is pre-trained for table understanding and then fine-tuned for column type annotation; TABBIE \cite{iida2021tabbie} separately processes the rows and columns of tables to give a better understanding of them; TCN \cite{wang2021tcn} suggests using both intra-table and inter-table information for column type prediction; Doduo \cite{suhara2022annotating} predicts all the columns of a table together by feeding the whole table to BERT; KGLink \cite{10597678} augments BERT with Wikidata to improve type predictions. REVEAL \cite{ding2025retrieve} selectively chooses the most relevant columns to the target column(s) before feeding them to BERT for prediction. These models rely on expensive fine-tuning of PLMs such as BERT: tables must be linearized into 1D sequences, context length is limited, and their semantic coverage lags behind modern LLMs.

\textbf{LLM-based prompting approaches.}
Recent work explores zero- and few-shot table annotation either by prompting general-purpose LLMs~\cite{korini2023column,korini2024column,zhang2024tablellama,kayali2024chorus} without supervised training, or by using synthetic supervision generated by LLMs~\cite{hoseinzade2026ztab,hoseinzadeefficient}, making them complementary to our supervised setting.

\section{Problem Definition}

\textit{Table annotation} includes three independent classification tasks: Column Type Annotation (CTA), Column Property Annotation (CPA), and Table Type Annotation (TTA), each involving different prediction targets and using a separate labeled dataset. Importantly, all tasks operate on tables without column headers, only the cell values are provided as input.

Formally, each task is defined over a dataset \( D \in \{D_{\text{CTA}}, D_{\text{CPA}}, D_{\text{TTA}}\} \), where both the tables and label sets can differ across tasks. Each table \( t = (c_1, c_2, \dots, c_n) \) consists of \( n \) columns and multiple rows of cell values. Each task follows a two-stage learning formulation. First, a feature extraction function \( \phi \) is learned that takes the full table as input and produces a sequence of structure-aware column embeddings:
\[
\psi = \phi(t) = \langle \psi_1, \psi_2, \dots, \psi_n \rangle, \quad \psi_i \in \mathbb{R}^d
\]
where \( \psi_i \) is the embedding for column \( c_i \). Then, a task-specific classifier \( f \) is learned to map the relevant subset of \( \psi \) to a label, depending on the task:

\begin{itemize}
    \item \textbf{CTA:} Given a column \( c_i \), predict its semantic type from a predefined, task-specific label set \( T_{\text{CTA}} \): 
    \[
    f_{\text{CTA}}(\psi_i) \rightarrow T_{\text{CTA}}
    \]
    
    \item \textbf{CPA:} Given a column pair \( (c_i, c_j) \), predict their relationship type from a predefined label set \( T_{\text{CPA}} \): 
    \[
    f_{\text{CPA}}(\psi_i, \psi_j) \rightarrow T_{\text{CPA}}
    \]
    
    \item \textbf{TTA:} Given a full table \( t \), predict its category from a predefined label set \( T_{\text{TTA}} \): 
    \[
    f_{\text{TTA}}(\psi_1, \dots, \psi_n) \rightarrow T_{\text{TTA}}
    \]
\end{itemize}

We treat each task independently, using separate datasets and training separate models. In this supervised setting, for each task we learn both the feature extractor \( \phi \) and the task-specific classifier \( f_{\text{task}} \) on its own labeled dataset \( D^{\text{train}}_{\text{CTA}} \), \( D^{\text{train}}_{\text{CPA}} \), and \( D^{\text{train}}_{\text{TTA}} \), respectively.

\section{TabEmb}

TabEmb operates with three components: a frozen large language model \( \mathcal{M}_{\text{LLM}} \) that produces semantic embeddings for each column, a graph neural network \( \mathcal{M}_{\text{GNN}} \) that refines these embeddings on the column graph into structure-aware column representations, and task-specific heads \( g_{\text{task}} \) that map the resulting representations to CTA, CPA, or TTA labels.

In the training phase, \( \mathcal{M}_{\text{LLM}} \) is kept frozen, and the parameters of \( \mathcal{M}_{\text{GNN}} \) and \( g_{\text{task}} \) are learned end-to-end from labeled tables for each task. In the prediction phase, the trained \( \mathcal{M}_{\text{GNN}} \) and \( g_{\text{task}} \) are used together with \( \mathcal{M}_{\text{LLM}} \) to annotate unseen tables. The following subsections describe these training and prediction procedures in detail.



\begin{algorithm}[t]
\small
\caption{Training Phase of TabEmb}
\label{alg:tabemb_training}
\begin{algorithmic}[1]
\REQUIRE Task \( \text{task} \in \{\text{CTA}, \text{CPA}, \text{TTA}\} \), Training dataset \( D^{\text{train}}_{\text{task}} \), frozen LLM \( \mathcal{M}_{\text{LLM}} \), trainable GNN \( \mathcal{M}_{\text{GNN}} \)

\ENSURE Trained GNN \( \mathcal{M}_{\text{GNN}} \), classifier \( g_{\text{task}} \)

\STATE \COMMENT{\textcolor{brown}{\textit{Module initialization}}}
\STATE \( g_{\text{task}} \leftarrow \textbf{TaskClassifier}(\text{task}) \)

\STATE \COMMENT{\textcolor{brown}{\textit{Semantic Embedding and Graph Construction}}}
\STATE \( \mathcal{G} \leftarrow \emptyset \)
\FOR{each \( (t, y) \in D^{\text{train}}_{\text{task}} \)}
    \STATE \( \psi^{(0)} \leftarrow \textbf{ColumnEmbedding}(t, \mathcal{M}_{\text{LLM}}) \)
    \STATE \( G_t \leftarrow \textbf{ConstructGraph}(t, \psi^{(0)}) \)
    \STATE Add \( (G_t, y) \) to \( \mathcal{G} \)
\ENDFOR


\STATE \COMMENT{\textcolor{brown}{\textit{Learning structure-aware embeddings and classifier}}}
\FOR{each training epoch}
    \FOR{each mini-batch \( \{(G_b, y_b)\}_{b=1}^{B} \subset \mathcal{G} \)}
        \STATE \( \mathcal{L} \leftarrow 0 \)
        \FOR{b = 1 to \(B\)}
            \STATE \( \psi \leftarrow \textbf{StructEmbedding}(G_b, \mathcal{M}_{\text{GNN}}) \)
            \STATE \( \hat{y} \leftarrow g_{\text{task}}(\psi) \)
            \STATE \( \mathcal{L} \leftarrow \mathcal{L} + \text{Loss}(\hat{y}, y_b) \)
        \ENDFOR
        \STATE Update \( \mathcal{M}_{\text{GNN}}, g_{\text{task}} \) using gradient descent on \( \mathcal{L} / B \)
    \ENDFOR
\ENDFOR


\RETURN \( \mathcal{M}_{\text{GNN}}, g_{\text{task}} \)
\end{algorithmic}
\end{algorithm}

\subsection{Training}

The training phase of TabEmb, described in Algorithm~\ref{alg:tabemb_training}, is run
\emph{separately} for each task \(\text{task} \in \{\text{CTA}, \text{CPA}, \text{TTA}\}\). For each training table \( (t, y) \in D^{\text{train}}_{\text{task}} \), the label \( y \) is defined according to the task: for CTA, it contains one semantic type per column; for CPA, it specifies a relation type for each column pair; and for TTA, it assigns a single type to the entire table. These labels guide the end to end supervised training of the GNN \( \mathcal{M}_{\text{GNN}} \) and task-specific classifier \( g_{\text{task}} \), which consists of three phases:

\textit{Module initialization.}
For a given task (CTA, CPA, or TTA), we instantiate the corresponding prediction head
\( g_{\text{task}} \) via \textbf{TaskClassifier(task)}, which selects a per-column,
per-pair, or pooled-table classifier with the appropriate input–output shape.
We also initialize a trainable graph neural network \( \mathcal{M}_{\text{GNN}} \) that refines LLM-based embeddings using structural relationships between table columns.

\textit{Semantic embedding and graph construction.}
For each labeled table \( (t, y) \in D^{\text{train}}_{\text{task}} \), we generate semantic embeddings for each column using the frozen LLM \( \mathcal{M}_{\text{LLM}} \). This is handled by \textbf{ColumnEmbedding($t(c_1, ..., c_n)$, \( \mathcal{M}_{\text{LLM}} \))}, which returns initial column embeddings
\( \psi^{(0)} = (\psi_1^{(0)}, \dots, \psi_n^{(0)}) \).
These embeddings are then used to construct a graph \( G_t \), where each column becomes a node initialized with its embedding, via \textbf{ConstructGraph($t(c_1, ..., c_n)$, \( \psi^{(0)} \))}. We precompute and store each labeled graph \( (G_t, y) \) in a graph pool \( \mathcal{G} \); frozen LLM is not used during subsequent gradient-based training.

\textit{Learning structure-aware embeddings and classifier.}
After preprocessing, we train the GNN and classifier end-to-end over the graph pool. We apply
\textbf{StructEmbedding(\( G_b, \mathcal{M}_{\text{GNN}} \))} to each graph \( G_b \) in the batch of labeled graphs \( \{(G_b, y_b)\}_{b=1}^{B} \subset \mathcal{G} \) to refine the column embeddings via message passing, obtaining structure-aware embeddings \( \psi \).
These are passed to the task-specific classifier \( g_{\text{task}} \) to generate predictions \( \hat{y} \).
We compute the loss \( \mathcal{L} = \text{Loss}(\hat{y}, y_b \)) over the batch of graphs and update the parameters of \( \mathcal{M}_{\text{GNN}} \) and \( g_{\text{task}} \) via gradient descent.

\vspace{1mm}
We define the core functions in Algorithm~\ref{alg:tabemb_training}:

\textbf{TaskClassifier(task):}
Returns the task-specific classifier \( g_{\text{task}} \): a per-column head for CTA, a per-pair head for CPA, or a per-table classifier for TTA. It is implemented as a simple linear mapping (single output projection, no hidden layers) from the learned embedding space to the task label space.

\textbf{ColumnEmbedding(t, \( \mathcal{M}_{\text{LLM}} \)):}  
For each column \( c_i \) in the table \( t = (c_1, \dots, c_n) \), we uniformly sample \( m \) non-null cell values, concatenate them into a text sequence, and feed it to the frozen LLM \( \mathcal{M}_{\text{LLM}} \). We then mean-pool the output token representations, i.e., take the average of all token embeddings, to obtain an initial column embedding \( \psi_i^{(0)} \in \mathbb{R}^d \), which we find more stable than using only the last token embedding. Repeating this for all columns yields the set of initial embeddings
\( \psi^{(0)} = (\psi_1^{(0)}, \dots, \psi_n^{(0)}) \).

\textbf{ConstructGraph(t, \( \psi^{(0)} \)):}  
Builds a fully connected undirected graph \( G_t = (V_t, E_t) \) for table \( t \), where each column \( c_i \) corresponds to a node \( v_i \in V_t \) initialized with its LLM-based embedding \( h_i^{(0)} = \psi_i^{(0)} \).
The graph includes both inter-column edges and self-loops to support information propagation and residual modeling. We use a fully connected column graph to avoid assuming prior knowledge about which column pairs are related. This allows the GNN to learn which neighboring columns are most informative and to weight their contributions accordingly. For very wide tables, where a fully connected graph may be costly, we can partition columns into coherent blocks (e.g., 10--15 columns via schema blocks) and construct a fully connected graph within each block.


\textbf{StructEmbedding(\( G_b, \mathcal{M}_{\text{GNN}} \)):}  
Applies a multi-layer GNN \( \mathcal{M}_{\text{GNN}} \) (e.g., a Graph Attention Network~\cite{velivckovic2017graph}) to \( G_b \), refining node representations through several rounds of message passing with residual connections. Starting from the initial node embeddings \( h_i^{(0)} \), the GNN layers iteratively update all nodes, and after \( S \) layers we obtain the final semantic–structure column embeddings
\(
\psi = (\psi_1, \dots, \psi_n),
\)
where each \( \psi_i \in \mathbb{R}^d \) is the refined embedding of column \( i \) in the table.

\begin{table*}[h]
\centering

\small
\begin{tabular}{l|ccc|cc|cccc}
\multicolumn{1}{c}{} & 
\multicolumn{3}{c}{\textbf{Single-column}} &  
\multicolumn{2}{c}{\textbf{Structure-augmented}} & 
\multicolumn{4}{c}{\textbf{Joint semantic-structure}} \\
 & Sherlock & RECA & ArcheType & SATO & GAIT & TURL & Doduo & REVEAL & \textbf{TabEmb (Ours)} \\ 
\toprule
CTA  & Yes & Yes & Yes & Yes & Yes & Yes & Yes & Yes & Yes \\
CPA  & Yes & No  & Yes & No  & No  & Yes & Yes & Yes & Yes\\
TTA  & Yes & No  & Yes & No  & No  & Yes & Yes & No  & Yes\\
\end{tabular}
\vspace{-2mm}
\caption{Different models categorized by how they model semantics and structure:
single-column (Sherlock~\cite{hul2019sherlock}, RECA~\cite{sun2023reca}, 
ArcheType~\cite{feuer2024archetype}),
structure-augmented (SATO~\cite{zhang2019sato}, GAIT~\cite{hoseinzade2024graph}),
and joint semantic-structure (TURL~\cite{deng2020turl}, Doduo~\cite{suhara2022annotating},
REVEAL~\cite{ding2025retrieve}), and our TabEmb.}
\label{tab:baseline}
\end{table*}

\begin{table*}
\centering
\small
\begin{tabular}{l|ccc|cc|cccc}
\multicolumn{1}{c}{} & 
\multicolumn{3}{c}{\textbf{Single-column}} &  
\multicolumn{2}{c}{\textbf{Structure-augmented}} & 
\multicolumn{4}{c}{\textbf{Joint semantic-structure}} \\

 & Sherlock 
 & RECA 
 & ArcheType
 & SATO 
 & GAIT
 & TURL 
 & Doduo 
 & REVEAL
 & \textbf{TabEmb (Ours)} \\ 
\toprule
CTA & 77.1 & 83.4 & 82.8 & 77.9 & 84.7 & 83.3 & 85.8 & 86.5 & \textbf{90.7} \\
CPA & 50.7 & N/A & 85.2 & N/A & N/A & 81.3 & 84.7 & 84.4 & \textbf{89.9} \\
TTA & 47.8 & N/A & 90.1 & N/A & N/A & 82.0 & 86.5 & N/A & \textbf{95.7} \\
\end{tabular}%
\vspace{-2mm}
\caption{Main results (micro-F1): baselines grouped by how they model semantics and structure. }
\label{tab:model_comparison}
\end{table*}

\begin{table}[t]
\centering
\small
\begin{tabular}{l|cc}
\textbf{Model}  & \textbf{AVG. train} & \textbf{AVG. test} \\
\toprule
Doduo          & 104.8   & 0.06 \\
ArcheType          & 585.9  & 0.35 \\
TabEmb             & \textbf{92.1} & 0.21 \\
\end{tabular}
\vspace{-2mm}
\caption{Average training time (minutes) and per-table test time (seconds) over all tasks and datasets.}
\label{tab:time_summary}
\end{table}

\subsection{Prediction}

At prediction time (Algorithm \ref{alg:tabemb_prediction}), TabEmb follows the same pipeline used during training. Given a new input table \( t = (c_1, \dots, c_n) \), the model first encodes each column independently using the frozen LLM to obtain initial embeddings \( \psi^{(0)} \). These embeddings are used to construct a fully connected graph \( G_t \), where each node represents a column and is initialized with its embedding.

The trained GNN \( \mathcal{M}_{\text{GNN}} \) is then applied to refine the embeddings through message passing over the graph, resulting in structure-aware representations \( \psi \). Finally, the task-specific classifier \( g_{\text{task}} \) consumes these refined embeddings to produce the output \( \hat{y} \), according to the task. 

\begin{algorithm}[h]
\small
\caption{Prediction Phase of TabEmb}
\label{alg:tabemb_prediction}
\begin{algorithmic}[1]
\REQUIRE Table \( t = (c_1, \dots, c_n) \), trained GNN \( \mathcal{M}_{\text{GNN}} \), trained classifier \( g_{\text{task}} \), frozen LLM \( \mathcal{M}_{\text{LLM}} \)
\ENSURE Predicted output \( \hat{y} \) for task \( \text{task} \in \{\text{CTA, CPA, TTA}\} \)

\STATE \( \psi^{(0)} \gets \textbf{ColumnEmbedding}(t, \mathcal{M}_{\text{LLM}}) \)
\STATE \( G_t \gets \textbf{ConstructGraph}(t, \psi^{(0)}) \)
\STATE \( \psi \gets \textbf{StructEmbedding}(G_t, \mathcal{M}_{\text{GNN}}) \)
\STATE \( \hat{y} \gets g_{\text{task}}(\psi) \)
\RETURN \( \hat{y} \)
\end{algorithmic}
\end{algorithm}

\section{Evaluation}

We evaluated TabEmb on the tasks of CTA, CPA, and TTA. All the experiments were run on a NVIDIA RTX 6000 Ada GPU.

\subsection{Method}

\textbf{\textbf{Datasets.}} 
We considered six datasets covering diverse domains and ontologies: T2D \cite{chen2019learning,t2d,korini2024column}, Wikitable \cite{deng2020turl}, Webtables \cite{suhara2022annotating, zhang2019sato,hoseinzade2024graph}, and the three SOTAB-V2 datasets: SOTAB\textsubscript{sch}, SOTAB\textsubscript{sch-s}, and SOTAB\textsubscript{dbp} \cite{sotab}. CTA is evaluated on all six datasets, CPA on all except Webtables, and TTA on the three SOTAB variants and T2D. More details on the choices of datasets 
are provided in Appendix \ref{appendix: dataset}. 
Across all datasets and tasks, we adopt the same train/test/validation splits or 5-fold cross-validation settings used in prior works.

\textbf{\textbf{Metric.}}
Following previous works \cite{feuer2024archetype, hoseinzadeefficient,hoseinzade2026ztab}, we evaluate the performance using \textit{micro-F1 score} collected on test tables, expressed in percentage. 

\textbf{Default LLM and GNN of TabEmb.}  We use a frozen Mistral-7B ~\cite{jiang2023mistral} (\( \mathcal{M}_{\text{LLM}} \)) via HuggingFace to compute column embeddings through sampling \(m{=}25\) non-null cell values per column. Specifically, we use the HuggingFace \emph{base} model \texttt{mistralai/Mistral-7B-v0.1} (not an Instruct variant).
A GAT \cite{velivckovic2017graph} with 2 hidden layers and 4 attention heads (\( \mathcal{M}_{\text{GNN}} \))
is implemented in DGL~\cite{wang2019deep} and trained with Adam (learning rate \(1\mathrm{e}{-3}\), weight decay \(5\mathrm{e}{-4}\)) for 100 epochs with a batch size of 256. We select the checkpoint with the best validation performance when a validation split is available, and otherwise use the final epoch. Unless otherwise stated, these settings
remain consistent for TabEmb across all experiments. Other types of LLMs and GNNs are investigated in Section \ref{sec: analysis}.

\textbf{Baselines.}
We compare TabEmb against eight \emph{supervised} baselines, grouped by how they model semantics and structure (Table~\ref{tab:baseline}). ArcheType is included only in its supervised regime. The zero-/few-shot LLM prompting methods~\cite{zhang2024tablellama,korini2023column,korini2024column,xiao2025cents} are not directly comparable to our supervised, content-only setting. 

\emph{Single-column.}
Sherlock, RECA, and ArcheType all support CTA. We adapt Sherlock and ArcheType by concatenating column pairs for CPA or all columns for TTA into a single input; RECA’s inter-table, single-column design does not extend beyond CTA.

\emph{Structure-augmented.}
SATO and GAIT are designed only for CTA, and their architectures do not naturally generalize to CPA or TTA.

\emph{Joint semantic–structure.}
TURL, Doduo, and REVEAL all support CTA and CPA; for TTA, we adapt TURL and Doduo via averaged column embeddings, while REVEAL’s context-selection per column mechanism does not extend cleanly. 



\begin{table*}[t]
\small
\centering
\resizebox{\textwidth}{!}{%
\begin{tabular}{l|ccc|cc|cccc}
\multicolumn{1}{c}{} & 
\multicolumn{3}{c}{\textbf{Single-column}} &  
\multicolumn{2}{c}{\textbf{Structure-augmented}} & 
\multicolumn{4}{c}{\textbf{Joint semantic-structure}} \\
\textbf{Dataset} 
  & Sherlock 
  & RECA 
  & ArcheType
  & SATO 
  & GAIT
  & TURL 
  & Doduo 
  & REVEAL
  & \textbf{TabEmb (Ours)} \\
\toprule
SOTAB\textsubscript{sch}    
  & 80.9 & 84.9 & 85.1 & 82.1 & 85.8 & 81.7 & 86.3 & 86.6 & \textbf{90.4} \\
SOTAB\textsubscript{sch-s}  
  & 72.6 & 81.9 & 83.0 & 74.2 & 83.1 & 75.4 & 81.1 & 83.1 & \textbf{87.5} \\
SOTAB\textsubscript{dbp}    
  & 77.5 & 83.1 & 83.6 & 79.0 & 84.7 & 76.7 & 85.2 & 86.1 & \textbf{91.0} \\
T2D                         
  & 67.7 & 86.5 & 88.0 & 70.7 & 85.1 & 94.0 & 91.1 & 91.7 & \textbf{95.5} \\
Wikitable                   
  & 75.8 & 74.2 & 76.7 & 76.0 & 75.5 & 78.9 & 75.2 & 75.8 & \textbf{82.9} \\
Webtables                   
  & 87.9 & 93.3 & 80.3 & 92.5 & 94.1 & 93.1 & 96.4 & 95.9 & \textbf{96.8} \\
\hline
Average                     
& 77.1 & 83.4 & 82.8 & 79.1 & 84.7 & 83.3 & 85.8 & 86.5 & \textbf{90.7} \\
\end{tabular}
}
\caption{CTA results: micro-F1 score comparison of TabEmb and baseline models.}
\label{table:column_type}
\vspace{3mm}

\centering
\begin{tabular}{l|cc|cccc}
\multicolumn{1}{c}{} & 
\multicolumn{2}{c}{\textbf{Single-column}} & 
\multicolumn{4}{c}{\textbf{Joint semantic-structure}} \\
\textbf{Dataset} 
  & Sherlock 
  & ArcheType
  & TURL 
  & Doduo 
  & REVEAL
  & \textbf{TabEmb (Ours)} \\
\toprule
SOTAB\textsubscript{sch}    
  & 67.6 & 88.6 & 83.3 & 90.2 & 90.5 & \textbf{91.7} \\
SOTAB\textsubscript{sch-s}  
  & 58.1 & 87.1 & 75.2 & 84.1 & 85.3 & \textbf{89.5} \\
SOTAB\textsubscript{dbp}    
  & 67.5 & 89.1 & 85.7 & 92.6 & 91.9 & \textbf{93.3} \\
T2D                         
  & 37.1 & 70.1 & 71.8 & 66.1 & 63.6 & \textbf{83.5} \\
Wikitable                   
  & 73.6 & 91.3 & 90.5 & 91.3 & 90.8 & \textbf{91.5} \\
\hline
Average                     
  & 50.7 & 85.2 & 81.3 & 84.7 & 84.4 & \textbf{89.9} \\
\end{tabular}
\caption{CPA results: micro-F1 score comparison of TabEmb and baseline models.}
\label{table:column_property}
\vspace{3mm}

\centering
\begin{tabular}{l|cc|ccc}
\multicolumn{1}{c}{} & 
\multicolumn{2}{c}{\textbf{Single-column}} & 
\multicolumn{3}{c}{\textbf{Joint semantic-structure}} \\
\textbf{Dataset} 
  & Sherlock 
  & ArcheType
  & TURL 
  & Doduo 
  & \textbf{TabEmb (Ours)} \\
\toprule
SOTAB\textsubscript{sch}    
  & 62.0 & 95.0 & 85.1 & 90.5 & \textbf{97.5} \\
SOTAB\textsubscript{sch-s}  
  & 57.6 & 91.7 & 81.7 & 86.4 & \textbf{96.5} \\
SOTAB\textsubscript{dbp}    
  & 49.2 & 94.2 & 82.5 & 86.1 & \textbf{97.4} \\
T2D                         
  & 23.1 & 79.4 & 78.4 & 83.0 & \textbf{91.5} \\
\hline
Average                     
  & 47.8 & 90.1 & 82.0 & 86.5 & \textbf{95.7} \\
\end{tabular}
\caption{TTA results: micro-F1 score comparison of TabEmb and baseline models.}
\label{table:table_type}
\end{table*}

\subsection{Main Results}

Table~\ref{tab:model_comparison} reports the average micro-F1 scores for CTA, CPA, and TTA across all applicable datasets. \textbf{Bold} indicates the best model in each task. Task-specific results are reported in the following Sections \ref{sec: cta}, \ref{sec: cpa}, \ref{sec: tta}. 

On CTA, TabEmb reaches 90.7 micro-F1, outperforming the strongest baseline REVEAL's 86.5 micro-F1 by +4.2 points.
On CPA, TabEmb attains 89.9 micro-F1, outperforming the strongest baseline ArcheType's 85.2 micro-F1 by +4.7 points.
On TTA, TabEmb attains 95.7 micro-F1, outperforming the strongest baseline ArcheType's 90.1 micro-F1 by +5.6 points.
Note that this "1-vs-all" comparison compares TabEmb with the best-performing baseline on each task.
The ``1-vs-1" comparison that compares TabEmb against one baseline at a time on all tasks would yield even larger gains.


Among \emph{single-column} models, ArcheType is the strongest baseline, but it still lags behind TabEmb, indicating that semantics alone are insufficient without explicit structural modeling.
GAIT is the best \emph{structure-augmented} model, yet TabEmb performs better because its embedding-level refinement learns richer inter-column relationships than GAIT’s prediction-level adjustment, which can only adjust classification scores based on other column's predictions.
Doduo and REVEAL are the strongest \emph{joint semantic–structure} baselines and are conceptually closest to TabEmb; however, their BERT-based encoders and tightly coupled text–structure modeling yield weaker semantic representations and less flexible structure modeling than TabEmb’s frozen LLM plus GNN design. 

Table~\ref{tab:time_summary} reports average training time (in minutes) and average test-time latency per table (in seconds) over all tasks and datasets for models that support all three tasks (excluding TURL and Sherlock due to low performance): Doduo, ArcheType, and TabEmb.
TabEmb, despite using a larger LLM backbone, has training time comparable to Doduo because it encodes columns once and then trains only a small GNN and task-specific heads. At test time, latency is dominated by the backbone LLM rather than the GNN or heads, so TabEmb is slower than Doduo but this cost is offset by its higher micro-F1 (+4.9, +5.2, +9.2 on CTA, CPA, and TTA). ArcheType is by far the slowest due to full-model LLM fine-tuning.


Next, we discuss task-specific results.

\subsection{CTA}
\label{sec: cta}
Table~\ref{table:column_type} presents CTA results across six datasets. TabEmb consistently outperforms all baselines on each dataset. In particular, following the top-down order of the table, TabEmb surpasses the strongest baseline of each dataset by +3.8\%, +4.4\%, +4.9\%, +1.5\%, +4.0\%, and +0.4\%, respectively.

Doduo and REVEAL are usually the strongest baselines, but on very small datasets like \textsc{T2D} (160 training tables) fine-tuning large models with many parameters (ArcheType, Doduo, REVEAL) on tabular inputs that differ from their pre-training objective is difficult and often unstable, so smaller models such as TURL perform relatively better. In contrast, TabEmb keeps the LLM frozen and trains only a GNN and a lightweight classification head, requiring far fewer parameters and achieving the best performance in this low-resource regime.

On the other hand, when there is enough labeled data to adapt models to tabular structure, LLM-based embeddings yield better generalization to unseen values than PLM-based embeddings, reducing reliance on large training sets. This is evident in the comparison between SOTAB\textsubscript{sch} and SOTAB\textsubscript{sch-s}, which share the same test and validation splits but differ in training size (44{,}769 vs.\ 10{,}631 tables, respectively). For instance, the BERT-based Doduo and REVEAL drop from 86.3 to 81.1 (\(-5.2\)) and from 86.6 to 83.1 (\(-3.5\)), while the LLM-based ArcheType and TabEmb go from 85.1 to 83.0 (\(-2.1\)) and from 90.4 to 87.5 (\(-2.9\)), respectively. Notably, TabEmb trained on the smaller SOTAB\textsubscript{sch-s} still outperforms the best baseline trained on the larger SOTAB\textsubscript{sch} (87.5 vs.\ 86.6), as its LLM-based column embeddings generalize better to unseen values.

\subsection{CPA}
\label{sec: cpa}
Table~\ref{table:column_property} presents the CPA results across five datasets. TabEmb consistently outperforms all baselines on each dataset. Again, following the top-down order of the table, TabEmb surpasses the strongest baseline of each dataset by +1.2\%, +2.4\%, +0.7\%, +11.7\%, and +0.2\%, respectively. Similar to CTA, all models perform worse on the low-resource T2D. 


Excluding T2D, the performance gap between TabEmb and strongest baselines (Doduo, ArcheType) is smaller on CPA than on CTA. CPA relies more on the row-wise value alignment between column pairs (e.g., subject–object patterns), which Doduo and ArcheType capture via signals from their row-level table access. TabEmb instead encodes whole column semantics with LLM embeddings and models inter-column dependencies via a GNN, prioritizing global semantic structure, which narrows the relative improvement on CPA.

\subsection{TTA} 
\label{sec: tta}
Table~\ref{table:table_type} reports TTA results on four datasets. TabEmb consistently outperforms all baselines, surpassing the strongest model on each dataset by +2.5\%, +4.8\%, +3.2\%, and +8.3\%, respectively. As with CTA and CPA, most methods report low performance on T2D due to limited training data.

Excluding T2D, ArcheType is the strongest baseline and outperforms Doduo on TTA: when whole table is considered as input, Doduo loses its structural advantage while ArcheType benefits from the richer semantics of an LLM over BERT. However, ArcheType still lags behind TabEmb. As a generative model, it may produce incorrect or non-canonical labels, whereas TabEmb is trained as a discriminative classifier 
to produce a fixed set of semantic type labels.

\subsection{Analysis} 
\label{sec: analysis}



We analyze TabEmb’s \emph{structure} and \emph{semantic} components via different GNN architectures and LLM backbones, respectively; we then examine generalization to rare classes and report sensitivity to the GNN depth and the number of sampled cell values per column.


\textbf{Structure Modeling.}
Table~\ref{tab:structure_aware_embedding} summarizes the impact of structural modeling in TabEmb. 
TabEmb (no GNN) is a purely semantic variant obtained by \textsc{StructEmbedding} (line~15 in Algorithm~\ref{alg:tabemb_training}) returning the initial GNN embeddings without any message passing, so the classifier operates directly on frozen LLM column embeddings.
In contrast, TabEmb (GAT), TabEmb (GCN), and TabEmb (GGNN) enable \textsc{StructEmbedding} with Graph Attention Networks (GAT)~\cite{velivckovic2017graph}, Graph Convolutional Networks (GCN)~\cite{kipf2016semi}, and Gated Graph Neural Networks (GGNN)~\cite{li2015gated}, respectively, performing joint semantic–structure modeling over the column graph.
All three structural variants improve over TabEmb (no GNN), confirming that joint semantic–structure modeling is beneficial and 
yields more effective embeddings.
Among them, TabEmb (GAT) performs better because its attention mechanism can focus on informative columns and down-weight noisy ones, whereas GCN and GGNN aggregate neighbors more uniformly. 

\begin{figure}[t]
\centering
\includegraphics[width=1\columnwidth]{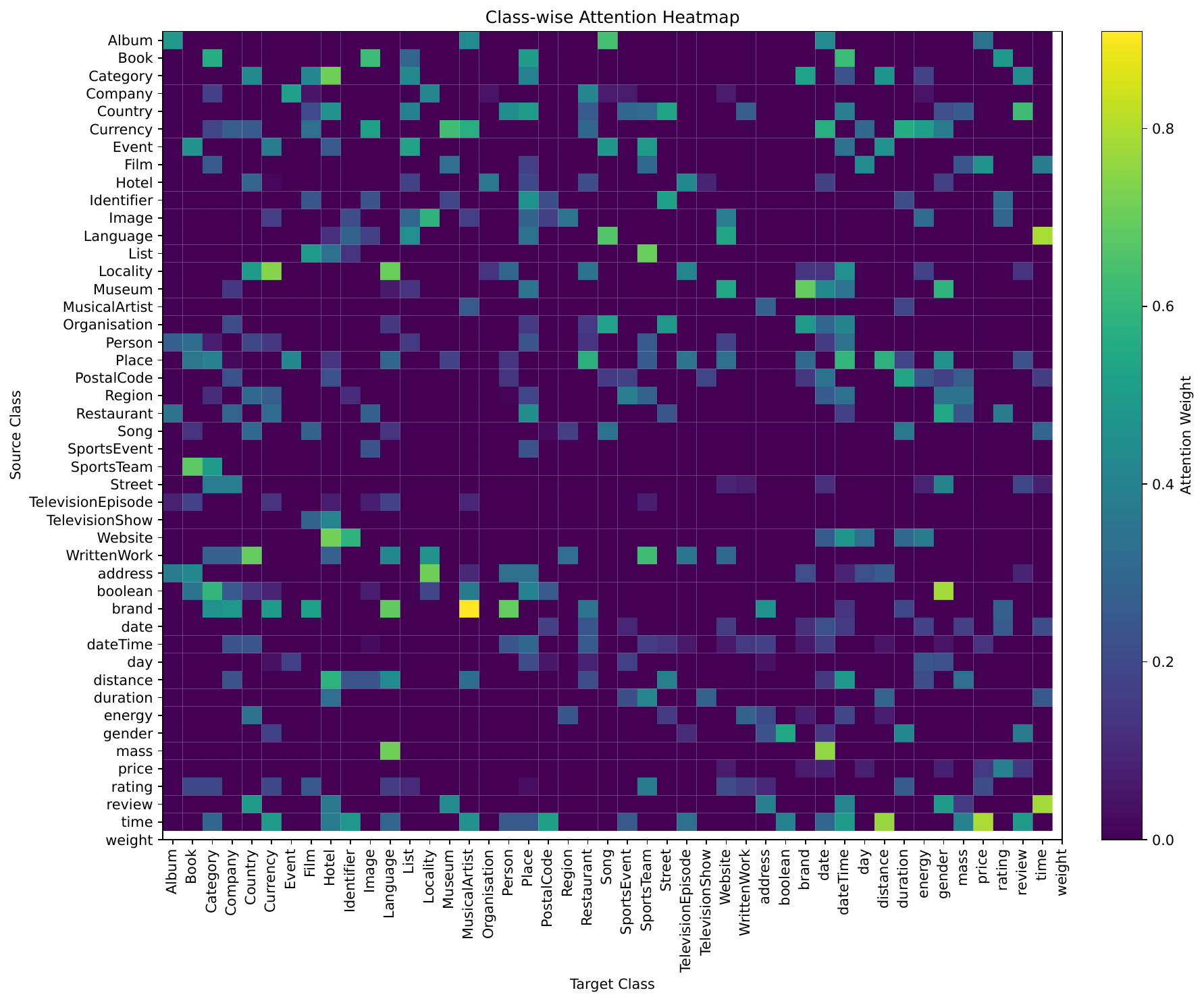} 
\vspace{-5mm}
\caption{class--class attention heatmap on SOTAB\textsubscript{dbp} (CTA), aggregated from GAT attention weights across tables.}
\label{fig:heatmap}
\end{figure}

Figure~\ref{fig:heatmap} shows a class--class attention heatmap on SOTAB\textsubscript{dbp} (CTA), aggregated from GAT attention weights across tables, indicating that the learned column interactions are highly non-uniform. Notable interactions often appear among semantically related groups, such as location-related types (\textit{address}, \textit{Place}, \textit{Street}, \textit{PostalCode}), temporal types (\textit{date}, \textit{dateTime}, \textit{time}, \textit{duration}), and measurement-related types (\textit{mass}, \textit{weight}, \textit{energy}), as well as music/media-related types (e.g., \textit{Album}, \textit{Song}, \textit{MusicalArtist}).

\begin{table}[t]
\small
\centering
\begin{tabular}{l|c}
\textbf{GNN used by TabEmb} & \textbf{GNN Avg.} \\
\midrule
No GNN & 88.3 \\
GAT                       & 92.1 \\
GCN                      & 90.5 \\
GGNN                     & 90.8 \\
\end{tabular}
\caption{Average micro-F1 scores (\%) over all tasks and datasets for each GNN-based  structural modeling. 
Per-task results are provided in Appendix \ref{appendix: structure} Table \ref{tab:structure_aware_embedding_full}.
}
\label{tab:structure_aware_embedding}
\end{table}


\textbf{Semantic Modeling.} Table \ref{tab:diff_llms} compares performance of seven \( \mathcal{M}_{\text{LLM}} \) models in TabEmb. Mistral achieves the highest average score (92.1), but the improvement over other LLMs is limited, suggesting that  there is not much difference between LLMs. In contrast, BERT performs worst with an overall average of 89.4, nearly 2.7 points lower than 
Mistral, indicating that older small PLMs like BERT capture insufficient semantic context.
Since TabEmb decouples semantic encoding from graph-based structure modeling, the LLM backbone can be replaced without changing the GNN or task heads (e.g., using embedding-specialized encoders such as e5/NV-Embed/Qwen or proprietary embedding APIs).

\begin{table}[h]
\small
\centering
\begin{tabular}{l|c}
\textbf{LLM used by TabEmb} & \textbf{LLM Avg.} \\
\midrule
Mistral (7B)      & 92.1 \\
Qwen (7B)         & 91.4 \\
Mixtral (45B)     & 91.9 \\
LLaMA-3 (8B)      & 91.3 \\
LLaMA-2 (7B)      & 91.6 \\
TinyLLaMA (1.1B)  & 91.3 \\
BERT (110M)       & 89.4 \\
\end{tabular}
\caption{Average micro-F1 scores (\%) over all tasks and datasets for each LLM-based semantic modeling. 
Per-task results are provided in Appendix \ref{appendix: semantic} Table \ref{tab:diff_llms_full}.
}
\label{tab:diff_llms}
\end{table}




\textbf{Generalization to Rare Classes.}\label{app:freq}
Table~\ref{tab:freq} reports a frequency-stratified analysis on SOTAB\textsubscript{dbp} for CTA/CPA/TTA. We split classes into three equally sized bins by frequency (High / Medium / Low) and report the average per-class F1 in each bin, comparing TabEmb against Doduo (the strongest joint semantic--structure baseline applicable to all three tasks). TabEmb’s per-class F1 remains relatively stable across bins, indicating strong performance even on less frequent classes. Moreover, the performance gap between TabEmb and Doduo generally widens from High to Medium/Low, consistent with TabEmb’s frozen LLM backbone providing richer world knowledge than Doduo’s BERT encoder and thus better generalization to rare semantic types.



{\color{Red}
\begin{table}[t]
\centering
\small
\begin{tabular}{llccc}
\multicolumn{2}{c}{} & \multicolumn{3}{c}{\textbf{Frequency}} \\
\cmidrule(lr){3-5}
\textbf{Task} & \textbf{Model} & \textbf{High} & \textbf{Medium} & \textbf{Low} \\
\midrule
CTA & TabEmb & 90.2 & 92.8 & 91.2 \\
CTA & Doduo  & 84.6 & 80.2 & 87.8 \\
\midrule
CPA & TabEmb & 94.4 & 87.7 & 96.4 \\
CPA & Doduo  & 92.0 & 83.1 & 93.2 \\
\midrule
TTA & TabEmb & 97.8 & 95.7 & 92.7 \\
TTA & Doduo  & 86.4 & 82.3 & 77.1 \\
\end{tabular}
\caption{Frequency-stratified per-class F1 on SOTAB\textsubscript{dbp}  (CTA/CPA/TTA), comparing TabEmb vs.\ Doduo.}
\label{tab:freq}
\end{table}
}

\textbf{Sensitivity to \( \mathcal{M}_{\text{GNN}} \) depth.} Table~\ref{tab:gat_depth} reports TabEmb's sensitivity to \( \mathcal{M}_{\text{GNN}} \) (GAT) depth, showing only modest variation across 1--4 layers, consistent with the complete-graph setting where even one layer provides global context.

\textbf{Sensitivity to row number $m$.} We observe that the number of sampled cell values per column ($m$ in the \textbf{ColumnEmbedding} function) has only a small effect once the LLM sees enough representative values: using $m{=}25$ (default) yields 92.1 avg micro-F1 vs.\ 91.7 for $m{=}15$, consistent with the LLM backbone providing strong semantic generalization beyond a moderate number of samples.


{\color{Red}
\begin{table}[t]
\centering
\small
\begin{tabular}{c|cccc}
\textbf{\#GAT layers} & \textbf{1} & \textbf{2} & \textbf{3} & \textbf{4} \\
\midrule
Avg micro-F1 & 91.9 & 92.1 & 92.4 & 92.0 \\
\end{tabular}
\vspace{-2mm}
\caption{Sensitivity to \( \mathcal{M}_{\text{GNN}} \) (GAT) depth (average micro-F1 over all tasks and datasets).}
\label{tab:gat_depth}
\vspace{-3mm}
\end{table}
}

\section{Conclusion}
We presented TabEmb, a general and efficient framework for table annotation that combines frozen LLMs for semantic encoding with GNN learning for structure-aware refinement. To support the three core tasks of column type annotation, column property annotation, and table type annotation, TabEmb uses a shared LLM+GNN architecture with lightweight task-specific classifiers. This design avoids complicated prompt engineering and costly LLM fine-tuning while outperforming state-of-the-art supervised baselines across multiple benchmark datasets. Our analysis highlights how embedding design choices can impact performance, offering guidance for building effective and scalable models for structured data understanding.

\section*{Acknowledgment}
The work of Ke Wang is supported in part by a discovery grant from Natural Sciences and Engineering Research Council of Canada.

\section*{Limitations}
TabEmb takes longer inference time than the BERT-based Doduo (Table 3) due to incorporating large LLMs, though TabEmb improves Doduo on micro-F1 by +4.9, +5.2, +9.2 on CTA, CPA, and TTA. Therefore, Doduo remains highly competitive in the scenario where inference time outweighs accuracy, whereas TabEmb will be the choice when accuracy is crucial. In addition, TabEmb currently constructs a fully connected column graph; for very wide tables, the quadratic number of inter-column edges can make graph computation expensive. While partitioning a table’s columns into several coherent blocks and constructing a fully connected graph within each block is a practical option, domain knowledge (e.g., schema constraints) can also be used to decide which column pairs should be connected by edges, introducing effective sparsity and simplifying the graph.


\bibliography{custom}

\appendix

\section{Dataset Details}
\label{appendix: dataset}

\begin{table}[h]
\centering
\small
\begin{tabular}{l|l|ccc}
\textbf{Dataset} & \textbf{Task} & \textbf{\#Class} & \textbf{\#Train} & \textbf{\#Test} \\
\toprule
\multirow{3}{*}{SOTAB\textsubscript{sch}} 
  & CTA & 82  & 44769 & 609 \\
  & CPA & 108 & 29158 & 565 \\
  & TTA & 17  & 29158 & 565 \\
\midrule
\multirow{3}{*}{SOTAB\textsubscript{sch-s}} 
  & CTA & 82  & 10631 & 609 \\
  & CPA & 108 & 7430  & 565 \\
  & TTA & 17  & 7430  & 565 \\
\midrule
\multirow{3}{*}{SOTAB\textsubscript{dbp}} 
  & CTA & 46 & 37631 & 279 \\
  & CPA & 49 & 22905 & 311 \\
  & TTA & 17 & 22905 & 311 \\
\midrule
\multirow{3}{*}{T2D} 
  & CTA & 37 & 160 & 109 \\
  & CPA & 48 & 81  & 82 \\
  & TTA & 23 & 81  & 82 \\
\midrule
\multirow{3}{*}{Wikitable} 
  & CTA & 255 & 397098 & 4764 \\
  & CPA & 121 & 52943  & 1467 \\
  & TTA & N/A & N/A    & N/A \\
\midrule
\multirow{3}{*}{Webtables} 
  & CTA & 78 & \multicolumn{2}{c}{78733} \\
  & CPA & N/A & N/A  & N/A \\
  & TTA & N/A & N/A  & N/A \\
\bottomrule
\end{tabular}
\caption{Dataset statistics per task. For each dataset and task, we report the number of classes (\#Class), training tables (\#Train), and test tables (\#Test). ``N/A'' indicates that the task is not available for the dataset.}
\label{tab:dataset_summary}
\end{table}

We evaluate TabEmb on CPA, CTA and TTA tasks over six datasets. A summary of datasets used in this paper is provided in Table~\ref{tab:dataset_summary}. 
For CTA, all six datasets are used. For CPA, all datasets (except Webtables) that provide relation-level annotations are used. For TTA, all CPA-compatible datasets except Wikitable and Webtables are repurposed. 
In case of T2D, we used the version introduced in \cite{chen2019learning} for CTA, and the version introduced in \cite{korini2024column} for CPA and TTA, wherein the table type labels used for the TTA task are extracted from \cite{t2d}. 
All three SOTAB variants define 17 distinct table types, which we use as supervision for TTA. Since Wikitable is a multi-label dataset, and TabEmb as well as several baselines require a single label per instance, we select the first available label when multiple labels are present.

\section{Structural Modeling}
\label{appendix: structure}

This section includes the per-task results of the structural modeling analysis under Section \ref{sec: analysis}. Table \ref{tab:structure_aware_embedding_full} is a more detailed version of table \ref{tab:structure_aware_embedding}.

\begin{table}[t]
\small
\centering
\resizebox{\columnwidth}{!}{
\begin{tabular}{l|l|c|c}
\textbf{GNN used by TabEmb} & \textbf{Task} & \textbf{Task Avg.} & \textbf{Model Avg.} \\
\midrule
\multirow{3}{*}{No GNN}
& CTA & 87.50 & \multirow{3}{*}{88.29} \\
& CPA & 83.97 & \\
& TTA & 93.42 & \\
\midrule
\multirow{3}{*}{GAT}
& CTA & 90.67 & \multirow{3}{*}{92.09} \\
& CPA & 89.88 & \\
& TTA & 95.72 & \\
\midrule
\multirow{3}{*}{GCN}
& CTA & 89.59 & \multirow{3}{*}{90.48} \\
& CPA & 87.57 & \\
& TTA & 94.28 & \\
\midrule
\multirow{3}{*}{GGNN}
& CTA & 89.84 & \multirow{3}{*}{90.78} \\
& CPA & 88.06 & \\
& TTA & 94.45 & \\
\end{tabular}
}
\caption{Average micro-F1 scores (\%) for TabEmb variants across tasks.}
\label{tab:structure_aware_embedding_full}
\end{table}

\section{Semantic Modeling}
\label{appendix: semantic}

This section includes the per-task results of the semantic modeling analysis under Section \ref{sec: analysis}. Table \ref{tab:diff_llms_full} is a more detailed version of Table \ref{tab:diff_llms}.

\begin{table}[h]
\small
\centering

\begin{tabular}{l|l|c|c}
\textbf{LLM used by TabEmb} & \textbf{Task} & \textbf{Task Avg.} & \textbf{LLM Avg.} \\
\midrule
\multirow{3}{*}{Mistral (7B)}
& CTA & 90.67 & \multirow{3}{*}{92.09} \\
& CPA & 89.88 & \\
& TTA & 95.72 & \\
\midrule
\multirow{3}{*}{Qwen (7B)}
& CTA & 90.52 & \multirow{3}{*}{91.40} \\
& CPA & 88.48 & \\
& TTA & 95.20 & \\
\midrule
\multirow{3}{*}{Mixtral (45B)}
& CTA & 91.02 & \multirow{3}{*}{91.87} \\
& CPA & 88.92 & \\
& TTA & 95.67 & \\
\midrule
\multirow{3}{*}{LLaMA-3 (8B)}
& CTA & 90.43 & \multirow{3}{*}{91.28} \\
& CPA & 88.61 & \\
& TTA & 94.80 & \\
\midrule
\multirow{3}{*}{LLaMA-2 (7B)}
& CTA & 90.49 & \multirow{3}{*}{91.63} \\
& CPA & 89.00 & \\
& TTA & 95.41 & \\
\midrule
\multirow{3}{*}{TinyLLaMA (1.1B)}
& CTA & 89.50 & \multirow{3}{*}{91.29} \\
& CPA & 88.72 & \\
& TTA & 95.66 & \\
\midrule
\multirow{3}{*}{BERT (110M)}
& CTA & 87.60 & \multirow{3}{*}{89.42} \\
& CPA & 86.93 & \\
& TTA & 93.72 & \\
\end{tabular}
\caption{Average micro-F1 scores (\%) for each LLM and task.}
\label{tab:diff_llms_full}
\end{table}

\end{document}